\documentclass[]{spie}  
\usepackage{subcaption}
\usepackage{amsmath,amsfonts, amscd, amssymb}       
\usepackage{bbm}
\usepackage{booktabs}                                     
\usepackage{graphicx}                                     
\usepackage{wrapfig}                                      
\usepackage{cleveref}                                     
\usepackage{url}                                     
\usepackage{units}                                        
\usepackage{mathtools}                                    
\usepackage{accents}                                      %
\usepackage{tikz}                                         
\usetikzlibrary{arrows, positioning, fit}
\usepackage{enumerate}                                    
\usepackage{autonum}                                      
\usepackage{mathrsfs}                                     
\usepackage{cancel}                                       
\usepackage[super]{nth}                                   
\usepackage[sharp]{easylist}                              
\usepackage{listings}                                     
\usepackage{bbm}
\usepackage{esint}                      
\usepackage{bm}
\ListProperties(Progressive*=3ex)

\crefformat{figure}{Fig.\ #2#1#3}
\crefformat{equation}{Eq.\ (#2#1#3)}

\DeclarePairedDelimiter\paren{(}{)}                  
\DeclarePairedDelimiter\brac{[}{]}                  

\providecommand\ATitle[1]{e}

\makeatletter
\catcode`! 3
\providecommand{\im}{{\mathrm i}}
\providecommand{\e}{{\mathrm e}}

\def\Transpose #1{\romannumeral0\expandafter
                  \Mar@Transpose@a\romannumeral`^^@\Mar@DoOneRow #1\\!\\}

\def\Mar@DoOneRow #1\\{\Mar@DoOneRow@a {}#1&^^@&}%

\def\Mar@DoOneRow@a #1#2&{%
    \if^^@\detokenize{#2}\expandafter\@gobble\fi
    \Mar@DoOneRow@a {#1#2\\}%
}%

\def\Mar@Transpose@a #1#2\\{\ifx!#2\expandafter\Mar@FinishTranspose\fi
    \expandafter\Mar@Transpose@b\romannumeral`^^@\Mar@DoOneRow@a {}#2&^^@&#1}

\def\Mar@Transpose@b #1#2^^@\\{\Mar@Join {}#2^^@!#1}

\def\Mar@Join #1#2\\#3!#4\\%
   {\if^^@\detokenize{#3}\expandafter\Mar@EndJoin\fi
    \Mar@Join {#1#2&#4\\}#3!}%

\def\Mar@EndJoin\Mar@Join #1^^@!^^@\\{\Mar@Transpose@a {#1^^@\\}}

\def\Mar@FinishTranspose
    #1&^^@&#2\\^^@\\{ #2}

\catcode`! 12
\makeatother

\DeclarePairedDelimiterX\setbuild[2]{\{}{\}}{#1~ \big|~ #2} 
\providecommand{\inter}{\bigcap}                         

\DeclareFontFamily{U}{wncy}{}
\DeclareFontShape{U}{wncy}{m}{n}{<->wncyr10}{}
\DeclareSymbolFont{mcy}{U}{wncy}{m}{n}
\DeclareMathSymbol{\Sha}{\mathord}{mcy}{"58}

\providecommand{\mZ}{\mathbb{Z}}                         
\providecommand{\mR}{\mathbb{R}}                         

\providecommand{\mL}{\mathcal{L}}

\providecommand{\Lp}[2]{\mathcal{L}^{#1}(#2)}            

\providecommand{\deff}{\coloneqq} 

\providecommand{\define}{\coloneqq}

\providecommand{\dd}[1]{\mathrm{d}#1}

\providecommand{\integral}[4]{\int\limits_{\mathclap{#1}}^{\mathclap{#2}}#3\ \dd#4}

\providecommand{\s}[2]{\sum\limits_{\mathclap{#1}}^{\mathclap{#2}}}
\providecommand{\unioni}[3]{\bigcup\limits_{#1}^{#2} #3} 
\providecommand{\limit}[2]{\lim\limits_{#1\to #2}}

\providecommand{\conv}{\star}

\providecommand{\fhat}[1]{\widehat{#1}} 

\providecommand{\conj}[1]{\overline{#1}}

\providecommand{\Id}{\mathbbm{1}}          

\DeclarePairedDelimiter\norm{\|}{\|}                  

\providecommand{\ip}[2]{\langle #1, #2\rangle} 

\newcommand{\glmnet}{\texttt{G\textsc{lmnet}}}
\newcommand{\scatnet}{\texttt{S\textsc{catnet}}}

\title{Underwater object classification using scattering transform of sonar signals}

\author[a]{Naoki Saito}
\author[a]{David S.\ Weber}

\affil[a]{UC Davis, One Shields Ave, Davis, CA, USA}

\authorinfo{Further author information: (Send correspondence to David S.\ Weber)\\ Naoki Saito: E-mail: saito@math.ucdavis.edu, \\\indent\ website: \url{https://www.math.ucdavis.edu/~saito/}\\
David S.\ Weber.: E-mail: dsweber@math.ucdavis.edu,\\ \indent \ website: \url{https://dsweber2.wordpress.com/}}

\begin{document}
\maketitle

\begin{abstract}
    In this paper, we apply the scattering transform (ST)---a nonlinear map based off of a convolutional neural network (CNN)---to classification of underwater objects using sonar signals. The ST formalizes the observation that the filters learned by a CNN have wavelet-like structure. We achieve effective binary classification both on a real dataset of Unexploded Ordinance (UXOs), as well as synthetically generated examples. We also explore the effects on the waveforms with respect to changes in the object domain (e.g., translation, rotation, and acoustic impedance, etc.), and examine the consequences coming from theoretical results for the scattering transform. We show that the scattering transform is capable of excellent classification on both the synthetic and real problems, thanks to having more quasi-invariance properties that are well-suited to translation and rotation of the object. 
\end{abstract}


\section{INTRODUCTION}
\label{sec:intro}  

Deep neural networks, and convolutional neural networks in particular, have proven quite effective at discerning hierarchical patterns in large datasets \cite{LeCun2015-uc}. Some examples include image classification \cite{Krizhevsky2012-qr}, face recognition \cite{Sun2013-qy}, and speech recognition \cite{Dahl2013-vn}, among many others. Clearly, something is going very right in the design of CNNs. However, the principles that account for this success are still somewhat elusive\cite{Mallat2016-gp}, as is the construction of systems that work well with few examples.

The scattering transform was created to remedy these issues. In 2012, Stephane Mallat and Joan Bruna published both theoretical results \cite{Mallat2012-ka} and numerical implementations \cite{Bruna2011-it} tying together convolutional neural networks and wavelet theory. They demonstrated that scattering networks of wavelets and modulus nonlinearities, are translation invariant in the limit of infinite scale, and Lipschitz continuous under non-uniform translation, i.e. $T_\tau(f)(x)=f\paren[\big]{x-\tau(x)}$ for $\tau$ with bounded gradient. Numerically, they achieved state of the art on image and texture classification problems.

 More recent work from Wiatowski and B\"olcskei have generalized the Lipschitz continuity result from wavelet transforms to frames, and more importantly, established that increasing the \textit{depth} of the network also leads to translation invariant features \cite{Wiatowski2015-pg}. There have been a number of follow up papers, including a discrete version of Wiatowski's result
 \cite{Wiatowski2016-kt}, a related method on graphs \cite{Chen2014-np}, and a pseudo-inverse using phase retrieval \cite{Mallat2015-kv}. There have also been a number of papers using the scattering transform in such problems as fetal heart rate classification \cite{Chudacek2014-ha}, age estimation from face images \cite{Chang2015-sd}, and voice detection in the presence of transient noise \cite{Dov2014-hw}.

\begin{figure}[ht]
  {\centering
  \includegraphics[width=0.65\textwidth]{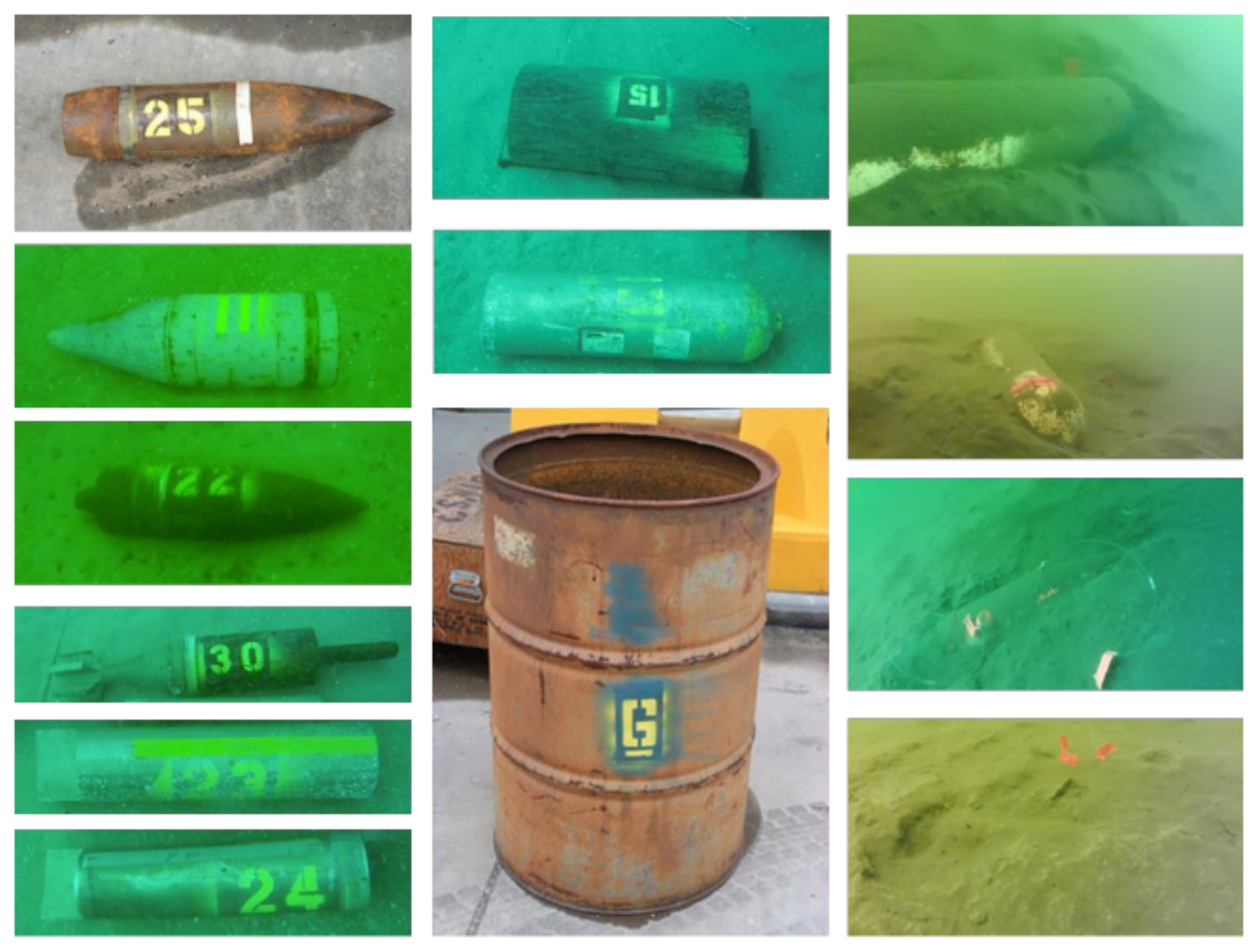}
  \caption{Example targets on shore and in the target environment\cite{geometryOfSonar}}\label{fig:stuff}}
\end{figure}

In this paper, we will apply the scattering transform of Mallat et al.\ to the problem of classification of underwater objects using sonar. This is motivated by previous work done by our group on the detection of unexploded ordinance (UXO) using synthetic aperture sonar (SAS) \cite{Marchand2007-ph}. From this, we use the BAYEX14 dataset \cite{geometryOfSonar}, which contains 14 different objects at various distances and rotations, partially submerged in a thin layer of mud on a sandy shallow ocean bed off Florida. \Cref{fig:stuff} shows examples of these objects. Additionally, we can generate synthetic examples using a fast solver for the Helmholtz equation in two regions with differing speed of sound, provided by Ian Sammis and James Bremer\cite{Bremer2012-cr,Bremer2013-zz}. We use this to examine more closely the dependence of both classification and the output of the scattering transform on both material properties and shape variations.

The approximate translation invariance and Lipschitz continuity under small diffeomorphisms of the scattering transform mean that for classes that are invariant under these transformations, members of the same class will be close together in the resulting space. So long as morphing via $f(t-\tau(t))$ from one class to another requires a $\tau$ with large derivative, then the classes will be well separated. Accordingly, we use a linear classifier on the output of the scattering transform. Additionally, because the scattering transform concentrates energy at coarser scales and wavelets in general encourage sparsity for smooth signals with singularities, we use a sparse version of logistic regression as our linear classifier, using LASSO \cite{Hastie2015-zf}. Specifically, we use a julia wrapper around \glmnet\cite{HastieTibshirani_2010}.

As a baseline classification scheme to compare the classification performance, we use the same linear classifier on the absolute value of the Fourier transform (AVFT) of the signal. One strong advantage of the absolute value of the Fourier transform is completely invariant to translations, and is the prototype of every linear filter which is invariant to translations \cite{OTSU,AMARI-PAT}. In addition to this invariant, the close ties between frequency and the speed of sound suggest that it should be sensitive to changes in the material. This will be examined in more depth in \Cref{sec:geom}.

\Cref{sec:scat} gives the basic set-up of a scattering transform, and approximate invariants from previous theoretical developments. These are increasing translation invariance with increasing depth, and Lipschitz stability to nonuniform time and frequency translation. \Cref{sec:SonarDetect} describes the setup used to generate both the synthetic and real datasets. \Cref{sec:results} gives the results of applying the ST and AVFT to the dataset to performing binary classification on shape and material in the synthetic case, and between UXO's and various other objects in the real case.

\section{Generalized Scattering Transform}\label{sec:scat}

\begin{figure}[ht]
		\centering
		\includegraphics[width=\textwidth]{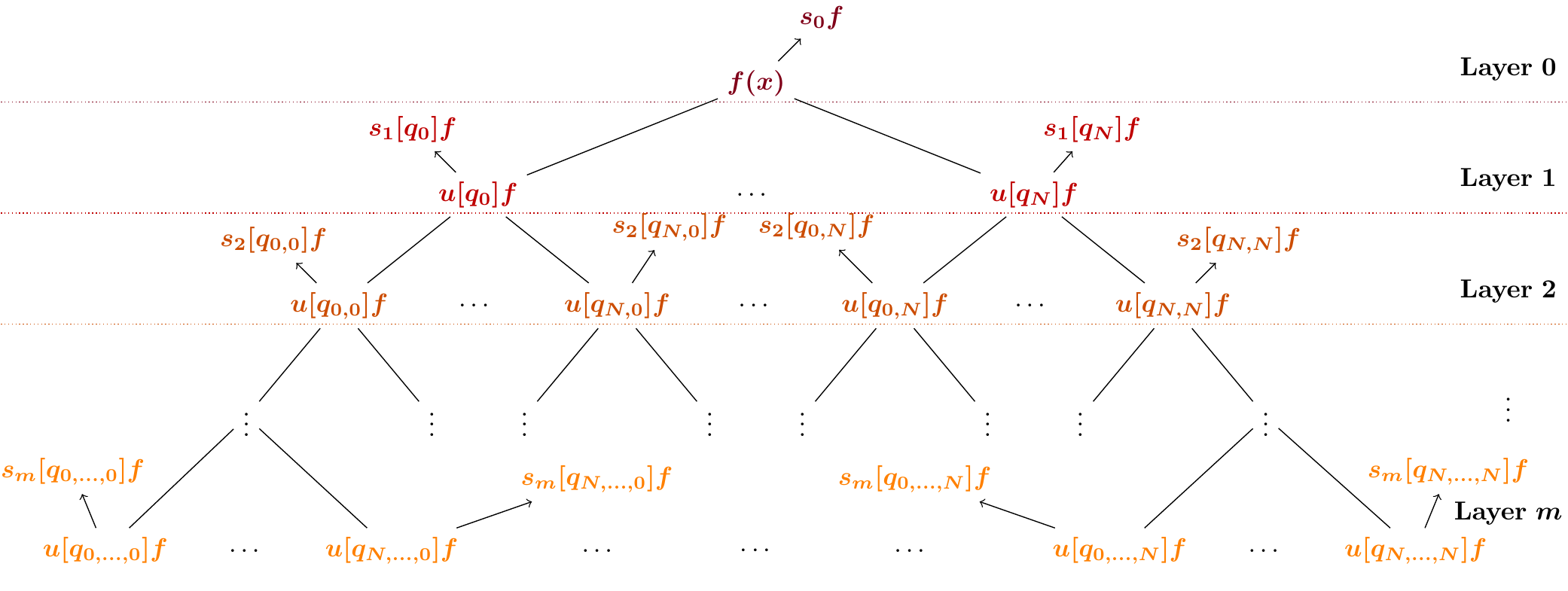}
		\caption{Generalized scattering transform. Here, $q_{k_n,\ldots,k_0}=(\lambda_{k_n}^{(n)},\ldots \lambda^{(1)}_{k_1})$ is an element of $\Lambda^n$, so $u_n[q_{k_m,\ldots,k_0}]$ is as in \cref{eq:recur}, and
		$s_n[q_{k_n,\ldots,k_0}]=\chi_{n}\conv u_n[q_{k_m,\ldots,k_0}]f$, i.e. an element of $\Phi_n$}\label{fig:gst}
\end{figure}

A generalized scattering network (hereafter referred to simply as a scattering network or ST)\cite{Wiatowski2015-pg,Mallat2012-ka}, has an architecture that is a continuous analog of a CNN. For a diagram, see \Cref{fig:gst}. First, at layer $m$, we start with a family of generators $\{g_{\lambda^{(m)}}\}_{\lambda^{(m)}\in\Lambda_m} \subset \mL^1(\mR^d)\inter \mL^2(\mR^d)$ for a translation invariant frame $\Psi_m = \{\psi
_{b,\lambda^{(m)}}\}_{b\in\mZ^d,\lambda^{(m)}\in\Lambda_m} = \{T_b I g_{\lambda^{(m)}} \}_{b\in\mZ^d,\lambda^{(m)}\in\Lambda_m}$ with frame bounds 
$a_m$ and $b_m$, that is
\begin{equation}
	a_m \norm{f}_2^2 \leq \s{\lambda^{(m)}\in\Lambda_m,b\in\mZ^d}{}\big|\ip{T_b C g_{\lambda^{(m)}}}{f}\big|^2
	\leq b_m\norm{f}_2^2,
\end{equation}
for all $f\in\mL^2(\mR^d)$. Here, $T_t[f](x)=f(x-t)$ is the translation operator, $C[f](x)=\conj{f(-x)}$ is the involution operator, and $\Lambda_m$ is some countable discrete index set, such as $\mZ$. This index set needs to tile the frequency plane in some way, for example by indexing scales and rotations. The frame atoms $g_{\lambda^{(m)}}$ correspond to the receptive fields found in each layer of a CNN. In addition, at layer $m$ we define a Lipschitz-continuous operator $M_m$ with bound $\gamma_m$ which satisfies $M_m f=0\Rightarrow f\equiv 0$. After these have been applied, the result is sub-sampled at a rate $r_m\geq 1$.
Finally, the output at each layer is then generated by averaging with a specific atom $\chi_m = g_{\lambda_*^{(m)}}$, which is removed from the set of frame atoms.

The scattering transform of Mallat\cite{Mallat2012-ka} for $\mR^d$ corresponds to choosing $\Lambda_m=\{2^{ {j/Q_m}}h\}_{j>-J_m,h\in H_m}$ for some finite rotation group $H_m$, $g_{0,\Id} = \psi(x)$ for some mother wavelet $\psi$, $\chi_m=\phi$ its corresponding father wavelet, and $Q_m$ a quality factor. The Lipschitz nonlinearity is $\big|\cdot\big|$.
 However, it lacks a sub-sampling factor, so $r_m=1$. Explicitly, the generator corresponding to index $\lambda^{(m)}=(j,h)$ is $g_{\lambda^{(m)}} = 2^{d j/{Q_m}}\psi(2^{ j/{Q_m}}h^{-1}x)$.

To get from layer $m-1$ to layer $m$, we define the function $u_m:\Lambda_m\times \mL^2(\mR^d)\to \mL^2(\mR^d)$ by
\begin{equation}
	u_m[\lambda^{(m)}]\paren[\big]{f}(z) = M_m[f\conv g_{\lambda^{(m)}}]\big(r_m z\big).
\end{equation}
Using this to define the value at a layer $m$, we have a path coming along a path of indices $q\in\Lambda^m \deff \Lambda_m \times \ldots \times \Lambda_1$,
\begin{equation}
	u[q](f) = u_m[\lambda^{(m)}]u_{m-1}[\lambda^{(m-1)}]\cdots u_1[\lambda^{(1)}]f\label{eq:recur}.
\end{equation}
Choosing $d=1$ corresponds to audio signals, such as sonar. Unlike a CNN, in a Scattering transform each layer has output, including just the average $\chi_m\conv f$, which is layer zero. For $m>1$, the output is
\begin{equation}
		\Phi_m[f] \deff \bigg\{u[q]f\conv\chi_m\bigg\}_{q\in\Lambda^m},
\end{equation}
while for the entire scattering transform,
\begin{equation}
	\Phi[f] \deff \unioni{m=0}{\infty} \Phi_m[f].
\end{equation}

\subsection{Previous theory}
There are two additional conditions that restrict the various operators in a given layer simultaneously\cite{Wiatowski2015-pg}. The first is the \textit{weak admissibility condition}, which requires that the upper frame bound $b_m$ be sufficiently small compared to the sub-sampling factor and Lipschitz constants:
\begin{align}
	\max\bigg\{b_m,\frac{b_m\gamma_m^2}{r_m^d}\bigg\} \leq 1\label{eq:wac},
\end{align}
which can be achieved by scaling the $g_{\lambda^{(m)}}$.

The second is that the nonlinearities must commute with the translation operator, so $M_mT_t[f] = T_tM_m[f]$. Most nonlinearities used for CNN's are pointwise i.e., $M_m[f](x)=\rho_m(f(x))$, so they certainly commute with $T_t$. Given these constraints, we can now state the results of Wiatowski and B\"olcskei\cite{Wiatowski2015-pg} precisely; the first is that the resulting features $\Phi_m$ deform stably with respect to small frequency and space deformations:

 {\theorem\cite{Wiatowski2015-pg} \label{thm:shifts}For frequency shift $\omega\in C(\mR^d;\mR)$ and space shift $\tau\in C^1(\mR^d)$, define the operator $F_{\tau,\omega}[f](x)\deff \e^{2\pi \im\omega(x)}f\big(x-\tau(x)\big)$.
	If $\norm{D\tau}_\infty\leq \frac{1}{2d}$, then there exists a $C>0$ independent of the choice of parameters for
$\Phi$ s.t. for all $f\in \mL _a^2(\mR^d)$, 
\begin{align}
	\norm[\big]{\Phi[F_{\tau,\omega}f]-\Phi[f]}_2 \define \s{m=1}{\infty}~\s{q\in\Lambda^m}{}\norm[\big]{\chi_{m} \conv u[q](F_{\tau,\omega}f) - \chi_m \conv u[q]f}_2\leq C (r\norm{\tau}_\infty + \norm{\omega}_\infty)\norm{f}_2\label{eq:deffree},
\end{align}}
where $\mL^2_a(\mR^d)$ is the set of $\mL^2(\mR^d)$ functions whose Fourier transforms are band limited to $[-a,a]$. 
Mallat shows a similar bound for the specific case that $M_m=|\cdot|$ and $g_{\lambda^{(m)}}$ is generated by an admissible mother wavelet $\psi$ with a number of conditions \cite{Mallat2012-ka}.

Their next result, translation invariance that increases with depth, is distinct from the one shown by Mallat where translation invariance increases with resolution:
{\theorem\cite{Wiatowski2015-pg} \label{thm:tranInv}Given the conditions above, for $f\in\mL^2(\mR^d)$ the $m^{th}$ layer's features satisfy
	\begin{align}
		\Phi_m\brac[\big]{T_t f} = T_{\frac{t}{r_1\cdots r_m}}\brac[\big]{\Phi_m(f)}\label{eq:fracTrans}.
	\end{align}
	If there is also a global bound $K$ on the decay of the Fourier transforms of the output features $\chi_n$:
	\begin{align}
		|\widehat{\chi_m}(\omega)| \leq K,
	\end{align}
	then we have the stronger result
	\begin{align}
		\s{n=1}{m}\norm[\big]{\Phi_n[T_tf]-\Phi_n[f]}_2 \leq \frac{2\pi|t|K}{r_1\cdots r_m} \norm{f}_2.
	\end{align}
}


To compare with the result from Mallat\cite{Mallat2012-ka}, that says that for an admissible mother wavelet $\psi$, the scattering transform achieves perfect translation invariance as the lower bound on the scale goes to infinity:
\begin{align}
  \limit{J}{\infty}\norm{\Phi_J[T_tf] - \Phi_J[f]} = 0.
\end{align}


\section{Sonar Detection}\label{sec:SonarDetect}
\begin{figure}[ht]
\begin{subfigure}[t]{.33\textwidth}
    \includegraphics[width=\textwidth]{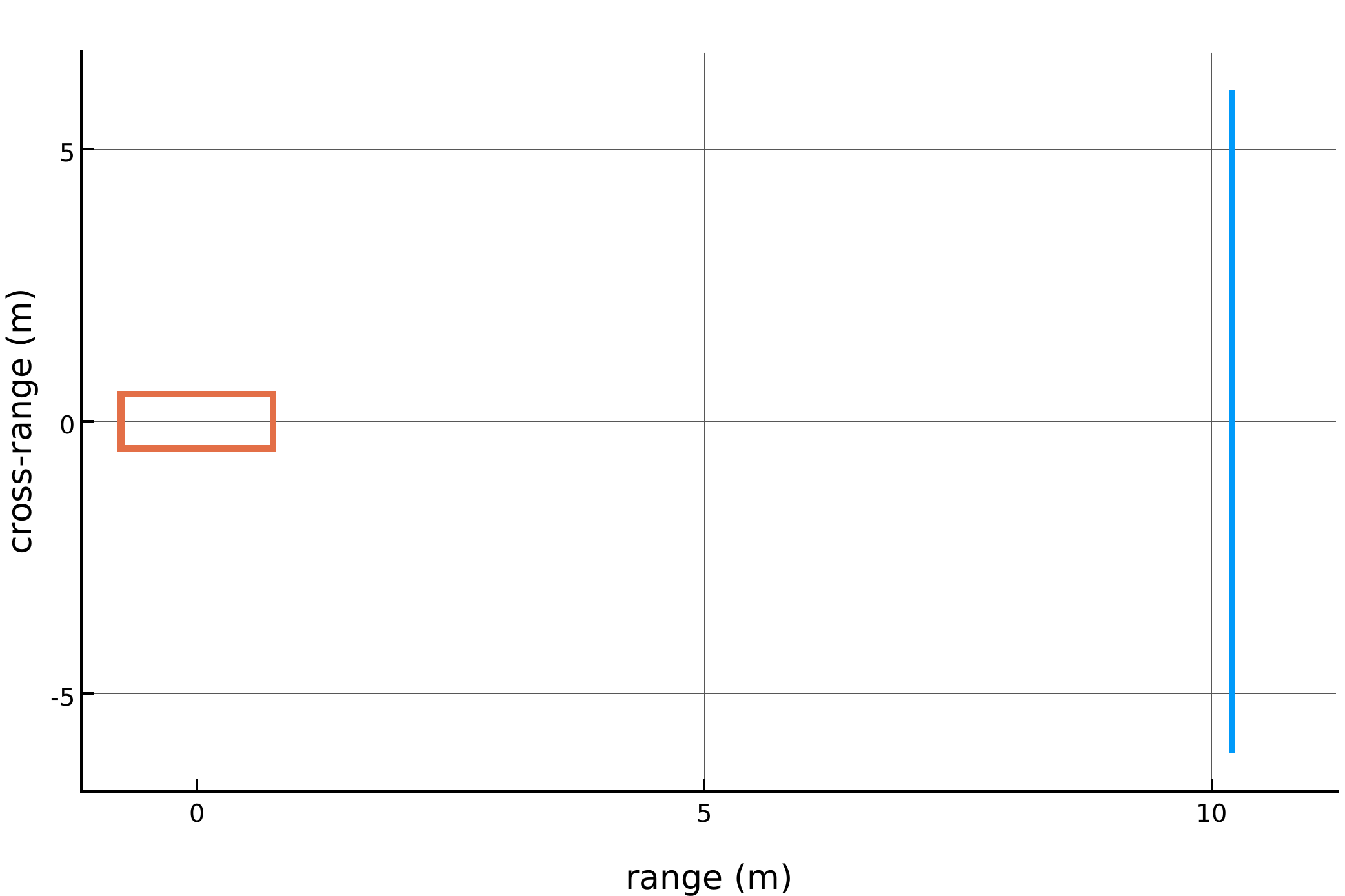}
\end{subfigure}%
\begin{subfigure}[t]{.33\textwidth}
    \includegraphics[width=\textwidth]{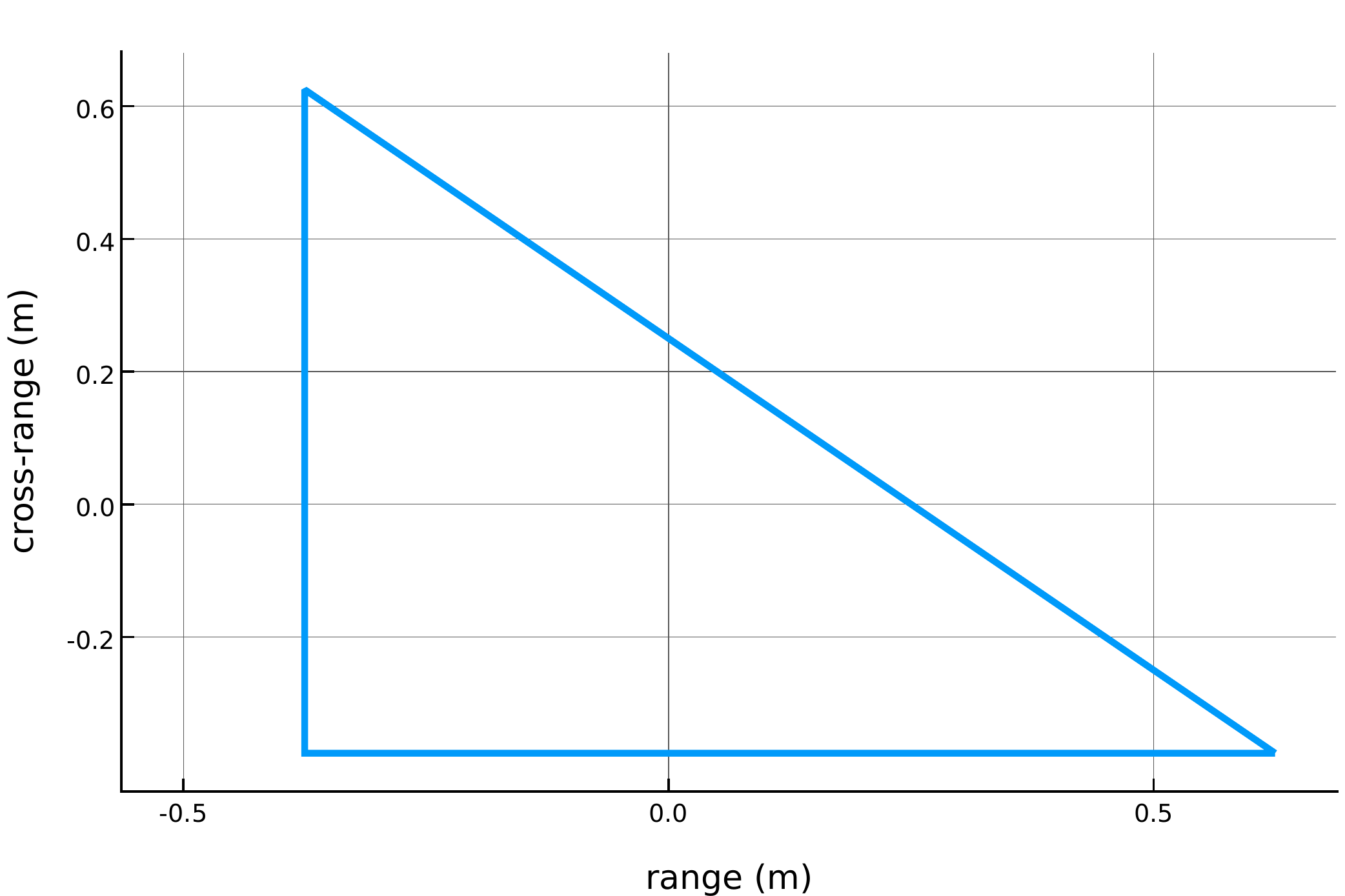}
\end{subfigure}
\begin{subfigure}[t]{.33\textwidth}
    \includegraphics[width=\textwidth]{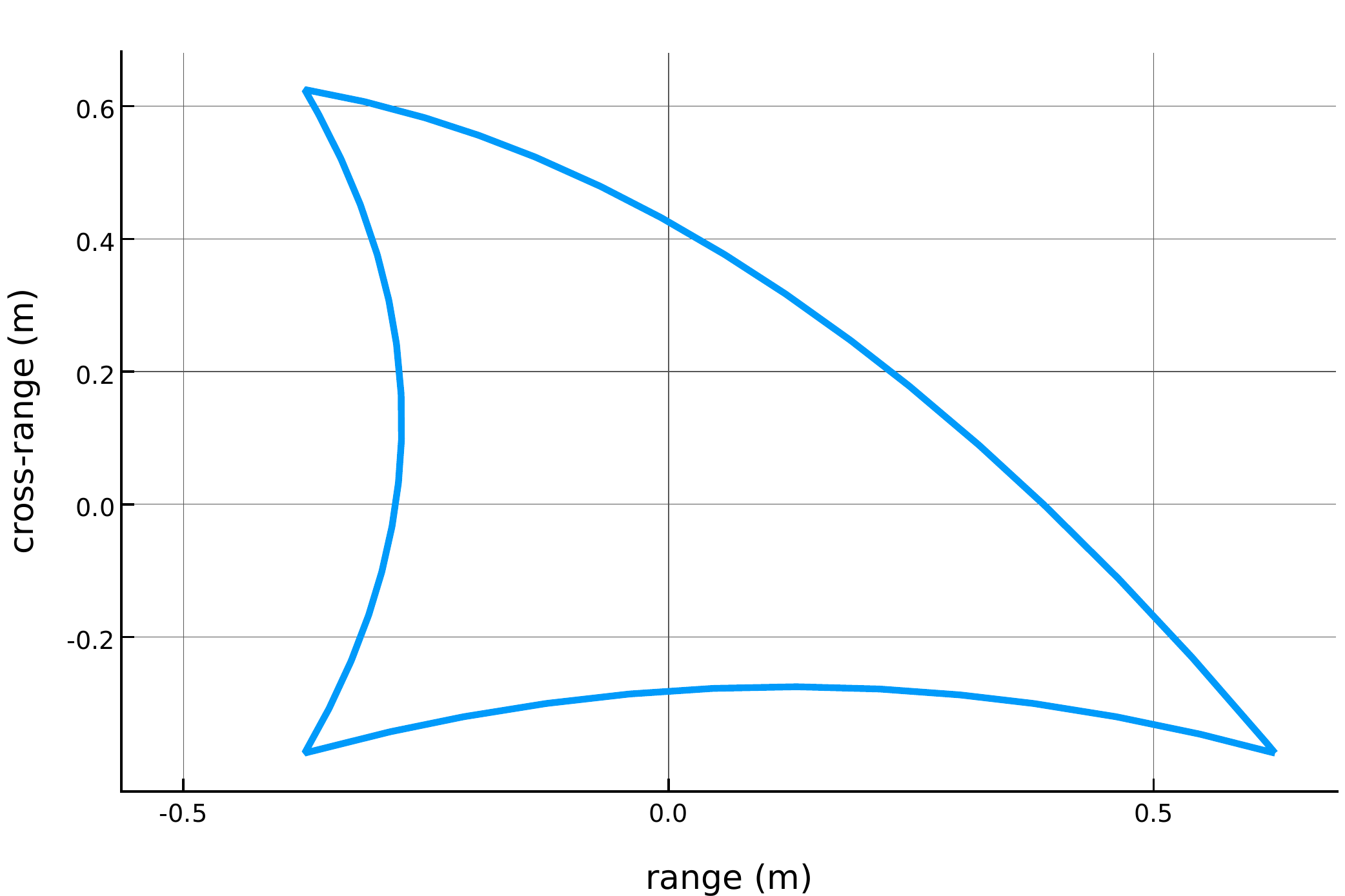}
\end{subfigure}
    \caption{The three shapes of $\Omega$ used in the synthetic setting. The rectangle on the left includes the observation rail without rotation. The range is approximately $10\textrm{m}$, while the observation rail itself is  $12\textrm{m}$ in total. The triangle has side lengths of $1\textrm{m}$, while the shark-fin is deformed from the same triangle. The rectangle has side lengths of $1\textrm{m}$ and $1.5\textrm{m}$.}
    \label{fig:geom}
\end{figure}

The problem that we will be investigating with the scattering transform techniques is the classification of objects partially buried on the sea floor using sonar data. This is motivated by using an unmanned underwater vehicle (UUV) equipped with sonar to detect unexploded ordinance. For this problem, we have both real and synthetic examples. The real examples consist of 14 partially buried objects at various distances and rotations, in a shallow mud layer on top of a sand ocean bed at a depth of $\sim$ 8m. BAYEX14 simulated the path of a UUV by stationing a sensor/emitter onto a rail, and pinging the field of objects was pinged at short intervals, see \Cref{fig:geom}. So for each rotation of the object relative to the rail, there is a 2D wavefield, where each signal corresponds to a location on the rail and the observation time.

 The synthetic examples come from considering the 2D Helmholtz equation in the regions with differing speed of sound\cite{Bremer2012-cr}:
\begin{align}
\begin{split}
\varDelta u + k_1^2 u = 0 \ \ \  & \mbox{ in } \Omega \\
\varDelta v + k_2^2 v = 0 \ \ \  & \mbox{ in } \Omega^c \\
u-v  = g \ \ \  & \mbox{ on } \partial\Omega \\
\partial_\nu u -\partial_\nu v = \partial_\nu g \ \ \   & \mbox{ on } \partial\Omega \\
\sqrt{|x|} \left( \partial_{|x|} -\im k_2 \right) v(x) & \to 0 \mbox{ as } |x|\to \infty,
\end{split}
\end{align}
  which gives the response to a sinusoidal signal with frequency $\omega$ on an object with $k_i={\omega/}{c_i}$, where $c_i$ is the speed of sound in the material, ranging from $343\textrm{m}/\textrm{s}$ in air, to $1503\textrm{m}/\textrm{s}$ in water, to $5100\textrm{m}/\textrm{s}$ in aluminum. It is worth noting that this is idealized in several ways: the model we use is 2D, rather than 3D, the material is modeled as a fluid with only one layer, instead of a solid with multiple different components, and there is no representation of the ocean floor itself.

\begin{figure}[ht]
	{\centering
		\includegraphics[width=.75\textwidth]{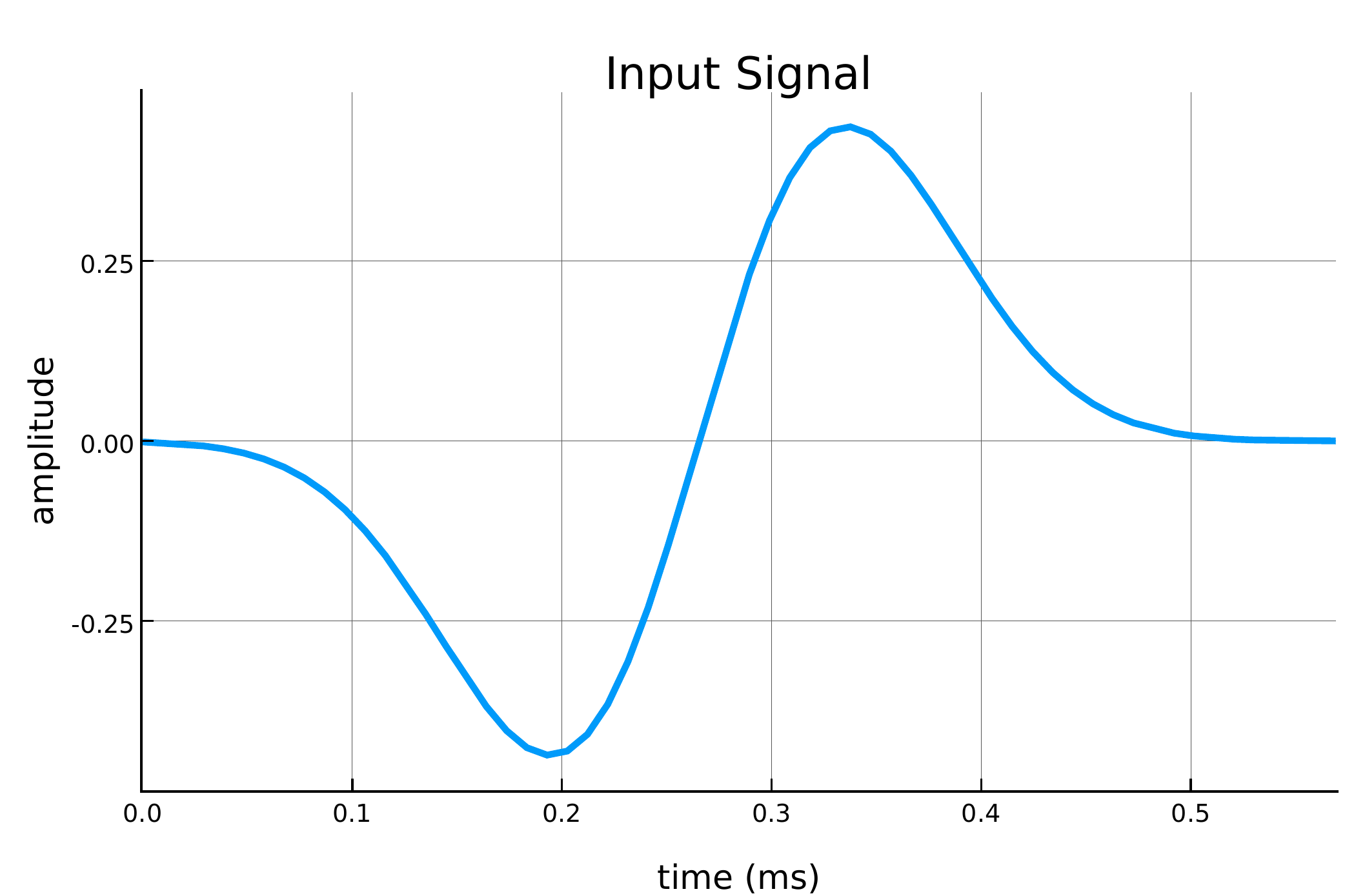}
		\caption{The non-zero portion of the source signal $s(t)$\label{fig:input}}
		}
\end{figure}

  The signals sent out by the UUVs are not pure sinusoids. One can approximate the response to multi-frequency signals (e.g. Gabor functions or chirps) by integrating across frequencies. We use a fast solver created by Ian Sammis and James Bremer to synthesize a set of examples, where we can more explicitly test the dependence of the scattering transform on the material properties (corresponding to changing the speed of sound) and geometry. The current dataset we use was created by Vincent Bodin, a former summer intern supervised by the first author. The input/source signal is shown in \Cref{fig:input}; zero padding is applied to each source signal to make the periodized version have approximately the same behavior as an input with finite support.

\begin{figure}[ht]
	\begin{center}
	  \includegraphics[width=0.48\textwidth]{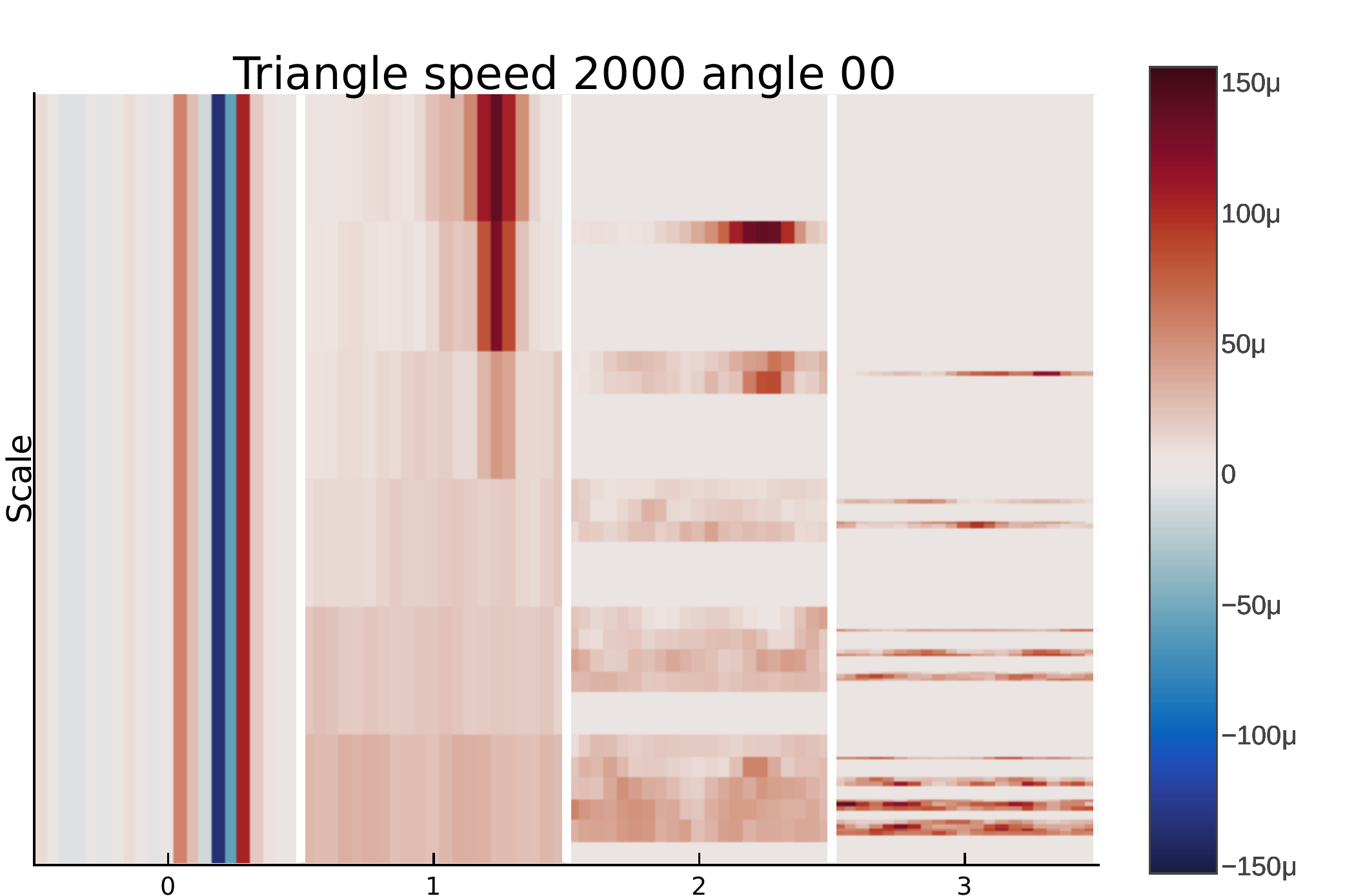}
	\end{center}
	\begin{flushleft}
	  \caption{The ST coefficients for the triangle object filled with a material whose speed of sound is $c_1=2000\textrm{m}/\textrm{s}$, while the ambient speed of sound is $c_2=1503\textrm{m}/\textrm{s}$. The x-axis refers to the depth, varying from $m=0$ to $m=3$. Scale is coarsest at the highest y-values, while within a given layer, the time increases horizontally. Note that only paths of increasing scale have been kept for computational reasons.}\label{fig:st}
	\end{flushleft}
\end{figure}
There are many design choices in testing the effectiveness of a classifier. As a baseline to compare against, we use the same \glmnet\ logistic classifier on the absolute value of the Fourier transform (AVFT), which is a simple classification technique that is translation invariant and sensitive to frequency shifts. To understand the generalization ability of the techniques, we split the data into two halves, one half training set and one half test set, uniformly at random 10 times, i.e. 10-fold cross validation. For the synthetic dataset, we also normalized the signals and added uniform Gaussian white noise so the SNR is 5dB.

For the synthetic data, there are two primary problems of interest. The first is determining the effects of varying shape on the scattering transform. An example of the 1D scattering transform for a triangle is in \Cref{fig:st}. Since the energy at each layer decays exponentially with layer index, layers 1-3 have been scaled to match the intensity of the first layer. Note that only the zeroth layer has negative values; this is because the nonlinearity used by the scattering transform is an absolute value. In the figure, one can clearly see a time concentrated portion of the signal in layers 0, 1, and 2.

For the real dataset, the problem of interest is somewhat more ambiguous. In addition to a set of UXO's and a set of arbitrary objects, there are some UXO replicas, not all of which are made of the same material. As we will see in the synthetic case, the difference in the speed of sound has a much clearer effect on classification accuracy than shape, so it is somewhat ambiguous how to treat these. We should expect that correctly classifying non-UXO's to be more difficult, since as a class they do not have much in common-- a SCUBA tank bears more resemblance to a UXO than it does to a rock.

\section{Geometric properties}\label{sec:geom}
Ideally the invariants discussed above would apply to transformations in the \textit{object} domain rather than to signals. But translations of the object (or equivalently, the observation rail), have a more complicated effect on the signal than simply translating the observation; at a minimum translating away from the object will cause a decay in signal amplitude. The changes in the object domain we seek to understand are changes in object material, translations and rotations of the object/rail, and changes in geometry. A classifier for this problem should be invariant to translation and rotation, but sensitive to the geometry of the object and the material.

 To understand more deeply how changes in the geometry affect the observed signals, we need to examine the solutions more closely. Let the input signal be fixed as $s(t)\in \Lp{1}{\mR}\inter\Lp{2}{\mR}$.
 The ideal reconstruction of the response to $s$, for a transmitter and receiver located at $\vec x\in\Omega^c$ is
\begin{equation}
  f(t,\vec{x}) = \integral{-\infty}{\infty}{\fhat{s}(\omega)v(\omega,\vec{x})\e^{-\im\omega \paren[\big]{t- P(\omega,\vec{x})}}}{\omega}\label{eq:recon}
\end{equation}
where $v$ is the solution, $\fhat{s}$ denotes the Fourier transform of $s$. To define the observation rail, first we define an unrotated observation, $\vec{x}_0(r)= (x,r)$ for $r\in [-y,y]$, so the rail has range $x=10\textrm{m}$ and length $2y=12\textrm{m}$. Then, a rotation of the object by an angle $-\theta$ is equivalent to rotating the rail by $\theta$, so define $\vec{x}_\theta(r)=R_\theta\vec{x}_0(r)$ where $R_\theta$ is the appropriate rotation matrix. In this setup, $x=10\textrm{m}$, and $y=6\textrm{m}$.

\begin{figure}[ht]
		{\centering
		\includegraphics[width=.75\textwidth]{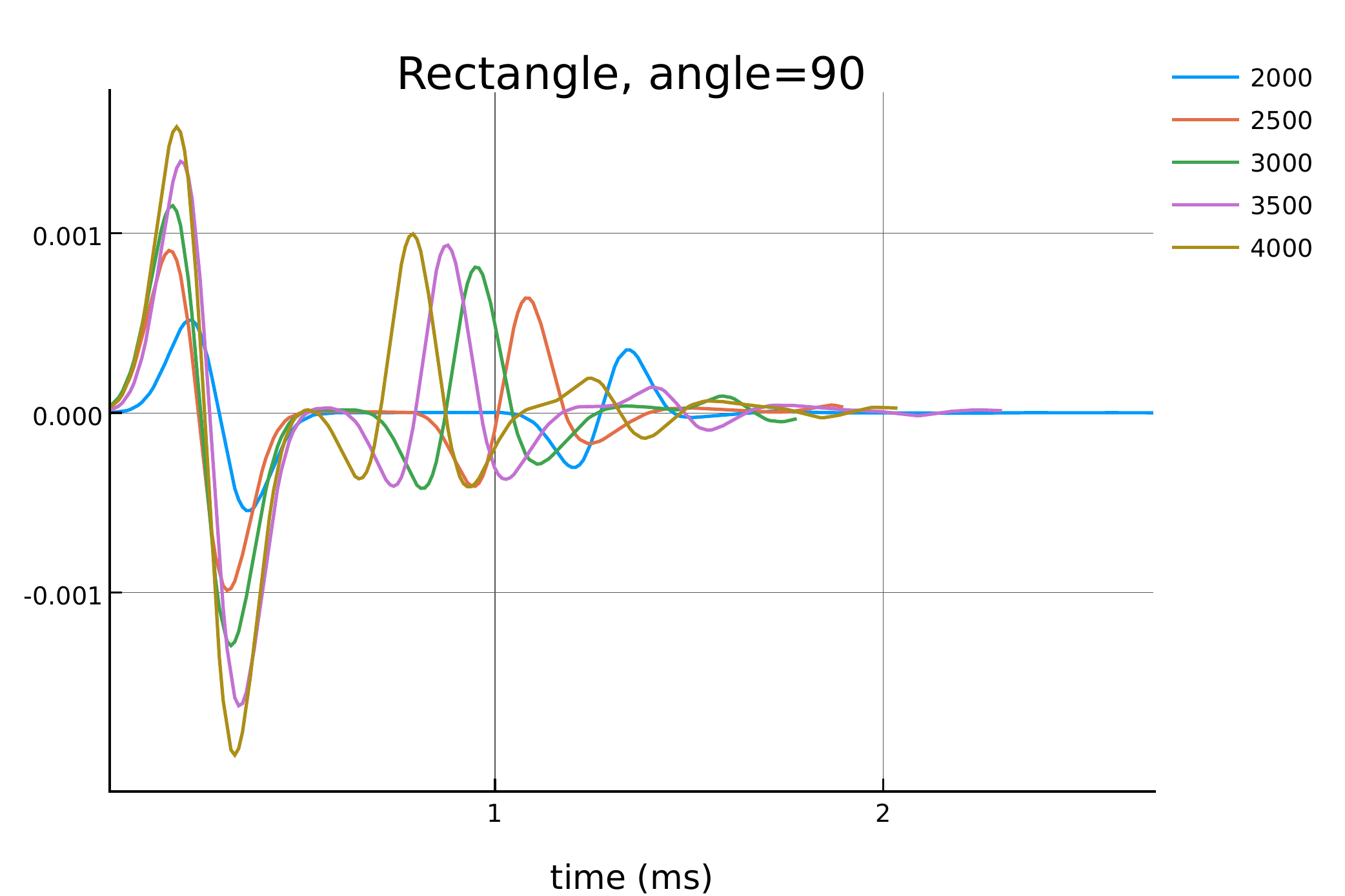}
	 \caption{Non-zero portion of the rectangle signal for varying speed of sound, when facing the long edge. \label{fig:rectLong}}
  }
\end{figure}
\subsection{Effects of the speed of sound}
To determine the behavior of a fixed location $x_\theta(r)$ as we change the speed of sound, we use common acoustic properties such as reflection coefficients and Snell's law. The crudest possible assumption that still gives meaningful results is that internal angles are irrelevant, and only refraction, reflection, and distance matters. Going from $\Omega^c$ with speed of sound $c_2$ to $\Omega$ with speed of sound $c_1$, the reflection coefficient is given by $V_{2,1} = \frac{Z_2-Z_1}{Z_1+Z_2}$, while the refraction coefficient is $W_{2,1} = 1-V_{2,1} = \frac{2Z_1}{Z_1+Z_2}$, where the impedance is $Z_i =\rho_i c_i$ ($\rho_i$ is the density of the material).
The distance from the center to a given point $x_\theta(r)$ on the line is just given by the Pythagorean theorem, $\sqrt{x^2+r^2}$. This means the initial peak occurs at $\frac{1}{c_2}\sqrt{x^2+r^2}$. If the input peak has magnitude $A_0$, then the first return peak should be approximately $A_1 = V_{2,1}\frac{A_0}{x^2+r^2} = \frac{(Z_2-Z_1)}{(x^2+r^2)(Z_2+Z_1)}A_0$; since $Z_2 > Z_1$ for most relevant examples, this is positive. Setting $\textrm{diam}(\Omega)=D$, the next peak can be approximated as $A_2 = W_{2,1}V_{1,2}W_{1,2}\frac{A_0}{(x^2+r^2)d} = \frac{4Z_2Z_1(Z_1-Z_2)} {(x^2+r^2)D(Z_1+Z_2)^3}A_0$; the sign of the second peak will flip.
The third peak is $A_3 = W_{2,1}V_{1,2}^2W_{1,2}\frac{A_0}{(x^2+r^2)D^2}$ and will flip sign again. Similarly $A_n= W_{2,1}V_{1,2}^n W_{1,2}\frac{A_0}{(x^2+r^2)D^n}$.

From this, we should expect that the decay from the first peak to the second peak, $\frac{4Z_1Z_2}{D(Z_1+Z_2)^2}$, is larger than that between any further consecutive peaks, $\frac{2|Z_1-Z_2|}{(Z_1+Z_2)D}$. One can see the sign flip clearly in \Cref{fig:rectLong} (take care that the input signal (\Cref{fig:input}) is a positive spike followed by a negative spike, so the second peak begins at $\sim 75 s$ for speed of sound $c_1\textrm{m}/\textrm{s}$).

For classification purposes, for a fixed angle and position on the rail, the reflection of every peak but the first depends on both $D$ and the speed of sound $c_1$. Further, the time between peaks should be ${D/}{c_1}$, so the scale of the solutions should provide a strong indicator of the speed of sound in the material. We will indeed see that the absolute value of a Fourier transform (AVFT), which has access to scale information, is reasonably effective at this problem. However for shapes with approximately the same diameter, such as the triangle and shark-fin, none of the above is useful for discrimination.

\subsection{Effects of rotation}
The discussion so far has almost completely ignored the internal geometry of $\Omega$. For a rectangle, when facing the longer side, this approximation is reasonable, as Snell's law matters when the internal angles depart significantly from $ \pi/ 2$. However, the problem of ray tracing in a region $\Omega$ is non-trivial in its own right, requiring averaging over all paths.

To avoid this issue, instead of trying to directly construct properties of the observations, we can derive how they will change under rotation and translation. We can use the far field approximation to do so \cite[Chpt.\ 4]{Etter2013}. Since $\textrm{diam}(\Omega)\approx 1$ and the center frequency of $s$ is $\omega_0 = 2500\textrm{Hz}$, we have that the range $10\textrm{m}\gg \frac{c_1}{\omega_0} \approx .6\textrm{m}$, the condition for far-field.
 Here we take the solution as approximately separable, so $v(\omega,\vec{x})\e^{\im \omega P(\omega,\vec{x})}\approx R(\omega,r)\Theta(\omega,\theta)$. In this representation, $R$ is a Bessel function of the first kind, and so to zero-th order is approximately $R(\omega,r)=J_0(k_2 r) \approx \frac{1}{\sqrt{k_2r}}\e^{\im k_2 r- \pi/ 4}$.


\begin{figure}[ht]
	{\centering
  \includegraphics[width=.55\textwidth]{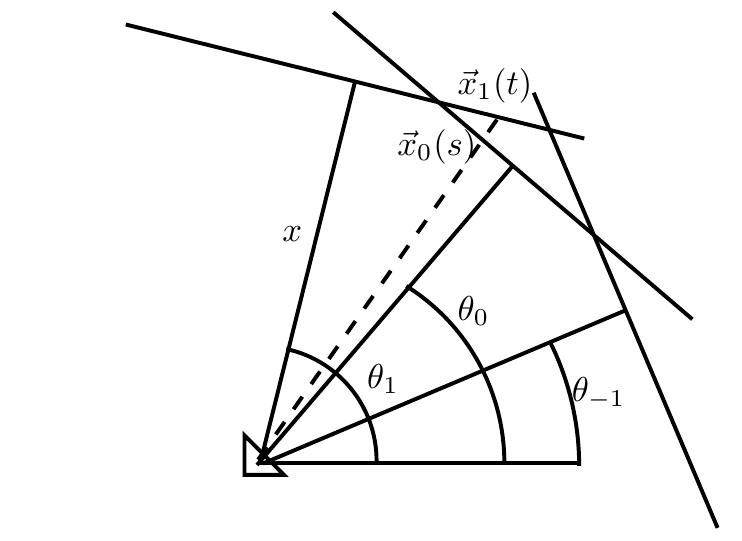}
  \caption{Composing a rail observation along $\vec{x}_0(q)$ from that of $\vec{x}_{-1}(p)$ and $\vec{x}_1(r)$.\label{fig:Rotation}}
  }
\end{figure}

Using this, we can examine the effect of rotation of the object (or the rail about the object). Suppose we know the solution at angles $\theta_{-1}$ and $\theta_{1}$, and we want to determine the solution at an angle $\theta_0$ between these. Every point in $\vec{x}_0(q)$ will have the same angle as a point on either $\vec{x}_1(r)$ or $\vec{x}_{-1}(p)$ (or possibly both) if $\theta_{-1}$ and $\theta_{1}$ are close enough that the paths cross, such as the case in \Cref{fig:Rotation}.
This happens when $\tan\paren[\big]{\frac 12 (\theta_1-\theta_{-1})}< y/x$, or in the synthetic dataset, $\theta_1-\theta_{-1}\leq  \pi/ 6$). Some geometry gives that the point $\vec{x}_0(q)$ has the same angle as $\vec{x}_1(r)$ if
\begin{equation}
  T(q) = -x\tan\paren[\big]{\theta_1-\theta_0-\arctan( q/x)}\label{eq:rotRad}
\end{equation}
for $q>0$; similar reasoning works for $q<0$ and $\vec{x}_{-1}$ with flipped signs. The distance changes from $\sqrt{x^2+q^2}$ to $\sqrt{x^2+T(q)^2}$, while the angle remains fixed, so plugging the zero-th order approximation to the Bessel function $J_0(k_2r)$ above into \cref{eq:recon},
\begin{align}
  f(r,\vec{x}_0(q)) &\approx \integral{-\infty}{\infty}{\fhat{s}(\omega) \frac{J_0\paren[\bigg]{k_2\sqrt{q^2+x^2}}}{J_0
		\paren[\bigg]{k_2\sqrt{T(q)^2+x^2}}} v\paren[\big]{\omega,\vec{x}_1\paren[\big]{T(q)}}\e^{-\im\omega \paren{t- P(\omega,\vec{x}_1(T(q)))}}}{\omega}\label{eq:full}\\
  f(t,\vec{x}_0(q)) &\approx \frac{\sqrt[4]{T(q)^2+x^2}}{\sqrt[4]{q^2+x^2}}
  \integral{-\infty}{\infty}{\fhat{s}(\omega) v\paren[\big]{\omega,\vec{x}_1\paren[\big]{T(q)}}\e^{-\im\omega \paren[\big]{t + h(q)- P(\omega,\vec{x}_1(T(q)))}}}{\omega}
\end{align}
where $h(q) =-\frac{1}{c_2}(\sqrt{q^2+x^2}-\sqrt{T(q)^2+x^2})$. So there are two effects on $f$ in the zero-th order approximation. The first is a phase shift by $h(q)$, under which both the AVFT and the ST are invariant. The second is a small amplitude modulation (for $|\theta_1-\theta_0| = \pi/ 6$, this ranges from .94 to 1.077). However, since the error in this approximation is $\paren[\big]{\frac{1}{k x}}^{ 1/2}\approx .24$, we should only roughly expect this to hold, since most of the signals have amplitudes on the order of $.001$.
In the full case of \cref{eq:full}, we have a ratio of Bessel functions $A(q,\omega)\e^{\im \kappa(q,\omega)} \define \frac{J_0\paren[\big]{k_2\sqrt{q^2+x^2}}}{J_0\paren[\big]{k_2\sqrt{T(q)^2+x^2}}}$ whose argument depends linearly on the frequency $\omega$. If $A(q,\omega)\leq 1$, then this can definitely be written in the form of a non-constant frequency shift $\omega(t)$ as in \Cref{thm:shifts}.

\subsection{Effects of translation}

Increasing the distance $x$ to the object from $x_1$ to $x_2$ is a similar transformation to rotation, since every point on the new rail corresponds to a point on the old rail, but with increased radius. If $r$ is the location on the new rail then the point with the same angle is just $T(r)=\frac{x_1}{x_2}r$; since $x_2>x_1$, this is smaller than $y$. Then we have the same sort of derivation as above, with a strictly linear function instead of \cref{eq:rotRad}.

For translation along a given rail, the dependence on $\Theta(\theta)$ is unavoidable. For a given point $x_\theta(r)$, the radius is simply $\sqrt{r^2+x^2}$, while the angle is $\varphi(r) = \theta + \arctan( r/x)$. We leave further discussion of the properties of $\Theta$ for our future publication.

\section{Results}\label{sec:results}

As noted before, for the linear classifier applied after the non-linear transform, we used sparse logistic regression as implemented in \glmnet\ \cite{Hastie2015-zf}. For the scattering transform, we used the \scatnet\ package as implemented by Mallat and his group \cite{Bruna2011-it}.

For the synthetic dataset, we compared three transforms. The first was the absolute value of the Fourier transform (AVFT). The second, hereafter referred to as the coarser ST, was a three layer scattering transform using Morlet Wavelets, with different quality factors rates in each layer: $Q_1 = 1, Q_2 = 1, Q_3 = 1$. Increasing quality factor corresponds to decreasing the rate of scaling of the mother wavelet, and thus gives more coefficients for the same frequency regime. The third, hereafter referred to as the finer ST, was another three layer scattering transform with increased quality in all 3 layers: $Q_1 = 8, Q_2 = 4, Q_3 = 4$. For each of these, we used 10-fold cross validation to check the generalization of our results.

As in \cref{sec:geom}, we investigated two problems: shape and material discrimination. For the shape discrimination, we compared the triangle and the shark-fin, with the material speed of sound fixed at $c_1=2000\textrm{m}/\textrm{s}$, while for the material discrimination, we fix $\Omega$ to be a triangle, and compared $c_1=2000\textrm{m}/\textrm{s}$ with $c_1=2500\textrm{m}/\textrm{s}$.
To do this comparison, we use the received operator characteristic (ROC) curve, which compares the trade off between false positives and true positives as we change the classification threshold; since it strictly concerns one class, it is insensitive to skewed class sizes\cite{Fawcett2006-qj}. A way of summarizing the ROC is the area under the curve (AUC), which simply integrates the total area underneath the curve; we use the trapezoidal approximation. This varies from .5 for random guessing to 1 for the ideal classifier, which doesn't misclassify.

\begin{figure}[ht]
 {\center
 \includegraphics[width=.5\textwidth]{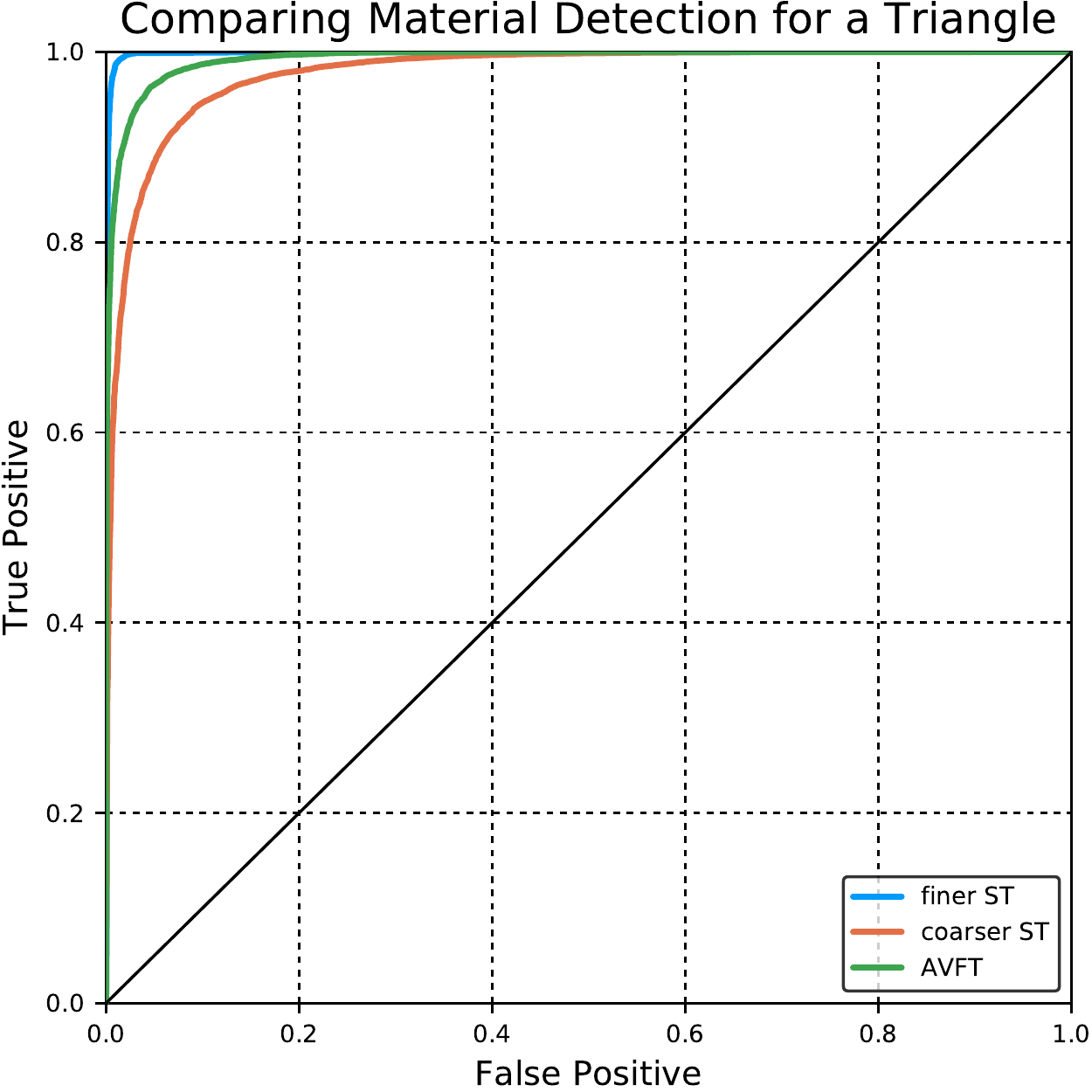}
 \caption{The ROC curve for detecting the material difference in a triangle, for speeds of sound $c_1=2000\textrm{m}/\textrm{s}$ and $c_1=2500\textrm{m}/\textrm{s}$. Note that the finer ST curve is an ideal classifier, completely in the upper left. The diagonal line is equivalent to random guessing.}
 \label{fig:ROCmat}}
\end{figure}
The results for material discrimination are in \Cref{fig:ROCmat}; the corresponding AUCs are $.992837$ for the AVFT, $.97778$ for the coarser ST, and $.99994$ for the finer ST. Fitting with the basic derivation in the geometry section, even the AVFT is capable of discriminating material effectively. Somewhat surprisingly, the coarser ST performed worse than the AVFT.  
This is likely because of insufficient frequency resolution, which the finer ST is able to achieve.

\begin{figure}[ht]
	{\center
	\includegraphics[width=.5\textwidth]{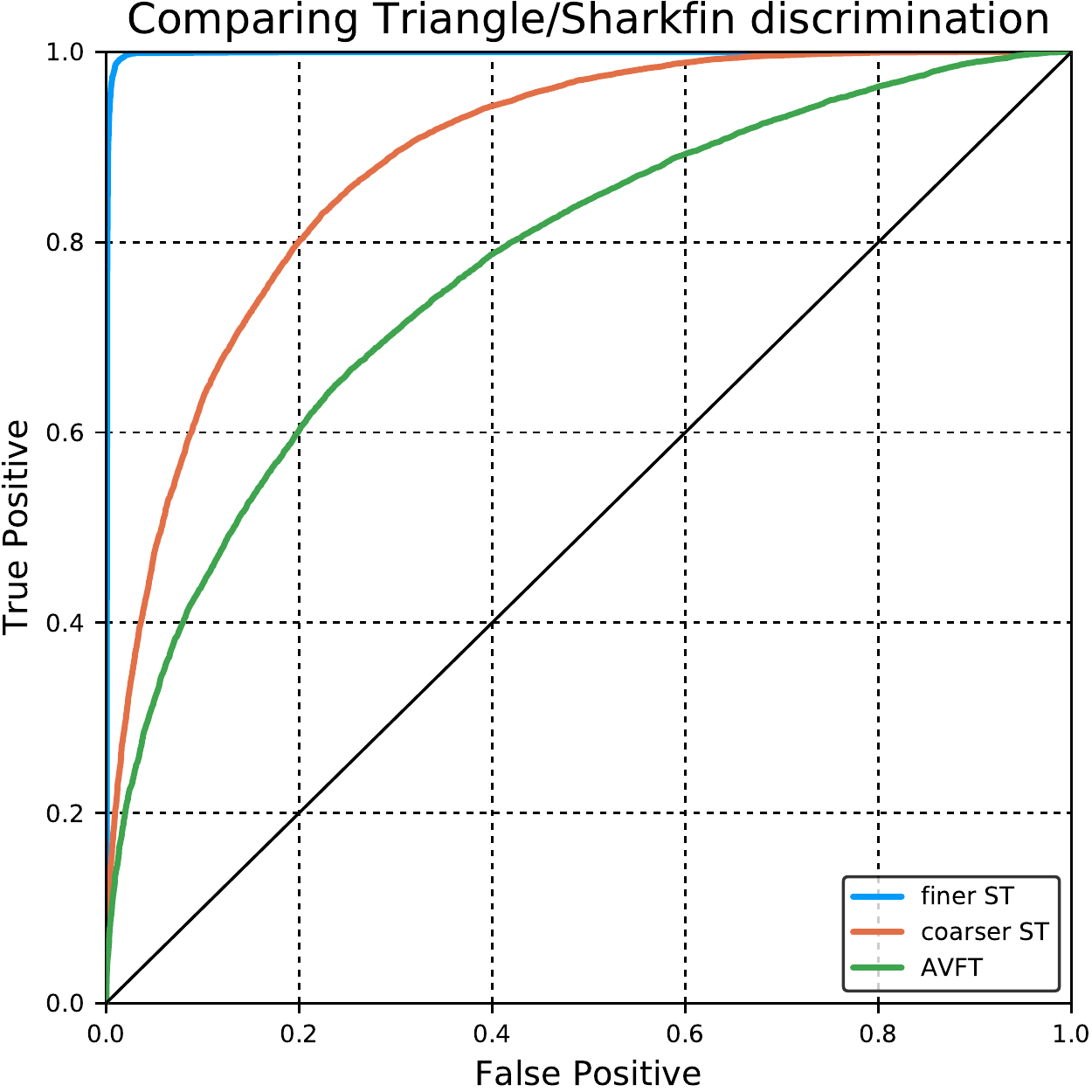}
	\caption{The ROC curve for discriminating a shark-fin from a triangle where both have a speed of sound fixed at $2000\textrm{m}/\textrm{s}$.}
	\label{fig:ROCshape}}
\end{figure}
The results in shape discrimination are more definitive in demonstrating the effectiveness of the ST, as seen in \Cref{fig:ROCshape}. The coarser scattering transform, with an AUC of .886, outperforms the AVFT with an AUC of $.775$. But the finer ST clearly outperforms both of these, with an AUC of $.998$, on par with the classification rates for the speed of sound problem.
\begin{table}
	\begin{center}
    \caption{The objects in each class for the real dataset\label{tab:objects}}
    \begin{tabular}{|c|c|}\hline
        UXO-like & Other Objects\\ \hline
        155mm Howitzer with collar & 55-gallon drum, filled with water\\
        152mm TP-T & rock\\
        155mm Howitzer w/o collar & 2ft aluminum pipe\\
        aluminum UXO replica & Scuba tank, water filled\\
        steel UXO replica & \\
        small Bullet&\\
        DEU trainer (mine-like object)&\\\hline
    \end{tabular}
   \end{center}
\end{table}

\begin{figure}[ht]
	{\center
	\includegraphics[width=.5\textwidth]{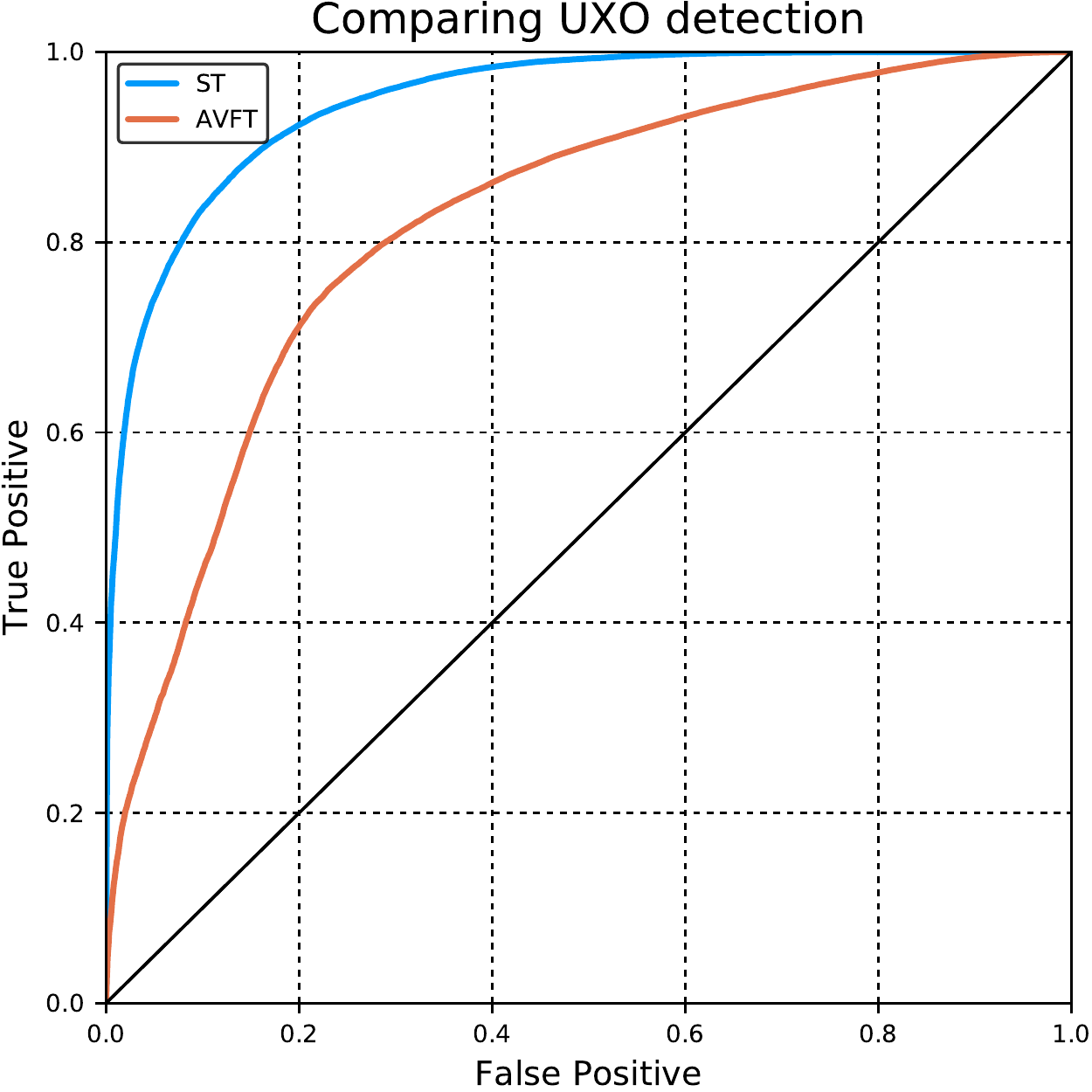}
	\caption{The ROC curve for detecting UXOs.}
	\label{fig:ROCreal}}
\end{figure}
For the real dataset, we compared two transforms, the AVFT and a two layer scattering transform with $Q_1=8,Q_2=1$. We have split the data into a set of objects that are either UXOs or replicas, and a set of the other objects in the dataset, as listed in \Cref{tab:objects}. In both classes, there are a variety of materials and shapes. Between classes, there are no similar shapes (as the shape is what determines if it is a replica rather than a UXO), but there are two with the same material (aluminum UXO replica vs aluminum pipe). The ST has an AUC of .9487, while the AVFT has an AUC of .8186. The AVFT actually did better on this problem than it had on the shape detection problem, suggesting that it is primarily the material properties of the UXOs that distinguish them.

\section{Summary}

In this paper, we have given initial results on understanding the problem of classifying sonar signals using the scattering transform. Using geometric arguments and the synthetic dataset, we have demonstrated that material detection is a considerably simpler problem than shape detection, and that the scattering transform is capable of solving both problems. We have given arguments for why the scattering transform may be suited to this problem by using previous theory, though it is difficult to accurately cast the transform in \cref{eq:full} in the form of a non-linear frequency modulation. Further, we have given numerical evidence that the scattering transform works well on this problem. This suggests that there may be additional operators for which the scattering transform has the Lipschitz stability of \cref{eq:deffree}, which better characterize the object domain.

\acknowledgments 
This research was supported in part by the grants from ONR N00014-12-1-0177, N00014-16-1-2255, and from NSF DMS-1418779. We would like to thank Frank Crosby and Julia Gazagnaire of Naval Surface Warfare Center, Panama City, FL, for providing us with the real BAYEX14 dataset. The work on the synthetic dataset is extended from the work of our former intern Vincent Bodin (now at Sinequa) and our former postdoctoral researcher Ian Sammis (now at Google) who first generated the dataset. James Bremer (UC Davis) and Ian Sammis developed the fast Helmholtz equation solver that we used. Finally, we would like to thank Stephane Mallat and his group (ENS, France) for their ST codes, Simon Kornblith (MIT) for the Julia wrapper of the \glmnet\ Fortran code, which was in turn developed by Jerome Friedman, Trevor Hastie, Rob Tibshirani and Noah Simon (Stanford).

\bibliography{bibliography} 
\bibliographystyle{spiebib} 
\end{document}